\documentclass{article} 
\usepackage{nips11submit_e,times}

\usepackage{algorithm}
\usepackage{algorithmic}
\usepackage{multirow}
\usepackage{times}
\usepackage{epsfig}
\usepackage{graphicx}
\usepackage{amsmath}
\usepackage{amssymb}
\usepackage{amsthm}
\usepackage{times}
\usepackage{graphicx}
\usepackage{psfrag}
\usepackage{url}
\usepackage{mathrsfs}
\usepackage{array}
\usepackage{bm}
\usepackage{subfigure}
\usepackage{subfigure}
\usepackage{wrapfig}
\usepackage[]{floatfig}

\usepackage{hyperref}

\newcommand{\figlabel}{Figure~}

\newcommand{\ImWidthErrorPlots}{0.45}
\newcommand{\ImWidthErrorPlotss}{0.45}

\newcommand{\ImWidthCostPlots}{0.247}
\newcommand{\ImWidthParamPlots}{0.42}
\newcommand{\ImWidthParamPlotss}{0.46}
\DeclareMathOperator*{\argmax}{argmax} 

\newenvironment{narrow}[2]{
  \begin{list}{}{
  \setlength{\topsep}{0pt}
  \setlength{\leftmargin}{#1}
  \setlength{\rightmargin}{#2}
  \setlength{\listparindent}{\parindent}
  \setlength{\itemindent}{\parindent}
  \setlength{\parsep}{\parskip}}
\item[]}{\end{list}}

\title{A Probabilistic Framework for Discriminative Dictionary Learning}

\author{
Bernard Ghanem and Narendra Ahuja
}

\nipsfinalcopy 

\begin{document}

\maketitle

\begin{abstract}
In this paper, we address the problem of discriminative dictionary learning (DDL), where sparse linear representation and classification are combined in a probabilistic framework. As such, a single discriminative dictionary and linear binary classifiers are learned jointly. By encoding sparse representation and discriminative classification models in a MAP setting, we propose a general optimization framework that allows for a data-driven tradeoff between faithful representation and accurate classification. As opposed to previous work, our learning methodology is capable of incorporating a diverse family of classification cost functions (including those used in popular boosting methods), while avoiding the need for involved optimization techniques. We show that DDL can be solved by a sequence of updates that make use of well-known and well-studied sparse coding and dictionary learning algorithms from the literature. To validate our DDL framework, we apply it to digit classification and face recognition and test it on standard benchmarks. 
\end{abstract}

\section{Introduction}\label{sec: intro}
Representation of signals as sparse linear combinations of a basis set is popular in the signal/image processing and machine learning communities. In this representation, a sample $\vec{\mathbf{y}}$ is described by a linear combination $\vec{\mathbf{x}}$ of a sparse number of columns in a dictionary $\mathbf{D}$, such that $\vec{\mathbf{y}}=\mathbf{D}\vec{\mathbf{x}}$. Significant theoretical progress has been made to determine the necessary and sufficient conditions, under which recovery of the sparsest representation using a predefined $\mathbf{D}$ is guaranteed \cite{Davenport2010,Wright2009b,Donoho2003}. Recent sparse coding methods achieve state-of-the-art results for various visual tasks, such as face recognition \cite{Wright2009}. Instead of minimizing the $\ell_0$ norm of $\vec{\mathbf{x}}$, these methods solve relaxed versions of the originally NP-hard problem, which we will refer to as \emph{traditional sparse coding} (TSC). However, it has been empirically shown that adapting $\mathbf{D}$ to underlying data can improve upon state-of-the-art techniques in various restoration and denoising tasks \cite{Elad2006,Mairal2008d}. This adaptation is made possible by solving a sparse matrix factorization problem, which we refer to as \emph{dictionary learning}. Learning $\mathbf{D}$ is done by alternating between TSC and dictionary updates \cite{KSVDalgo,MODalgo,Mairal2009,Jenatton2010}. For an overview of TSC, dictionary learning, and some of their applications, we refer the reader to \cite{Wright2010,Elad2010}.

In this paper, we address the problem of discriminative dictionary learning (DDL), where $\mathbf{D}$ is viewed as a linear mapping between the original data space and the space of sparse representations, whose dimensionality is usually higher. In DDL, we seek an optimal mapping that yields faithful sparse representation \emph{and} allows for maximal discriminability between labeled data. These two objectives are seldom complimentary and they tend to introduce conflicting goals in many cases, thus, classification can be viewed as a \emph{regularizer} for reliable representation and vice versa. From both viewpoints, this regularization is important to prevent overfitting to the labeled data. Therefore, instead of optimizing both objectives simultaneously, we seek joint optimization. In the case of sparse linear representation, the problem of DDL was recently introduced and developed in \cite{SDLNIPS2008,discriminative_reconstructive_hinge,Mairal2008c}, under the name \emph{supervised dictionary learning} (SDL). In this paper, we denote the problem as DDL instead of SDL, since DDL inherently includes the semi-supervised case. SDL is also addressed in a recent work on task-driven dictionary learning \cite{TaskDrivenDL2010}. The form of the optimization problem in SDL is shown in Eq. (\ref{eq: SDL}). The objective is a linear combination of a representation cost $e_R$ and a classification cost $e_C$ using data labels $\mathbf{L}$ and classifier parameters $\mathbf{W}$.


\begin{align}
\underset{\mathbf{X},\mathbf{D},\mathbf{W}}{\min} ~e_R\left(\mathbf{Y},\mathbf{X},\mathbf{D}\right)+\lambda e_C\left(\mathbf{X},\mathbf{W},\mathbf{L}\right) \label{eq: SDL}
\end{align}


\iffalse
\begin{figure}[ht]
\centering
\includegraphics[width=0.46\textwidth]{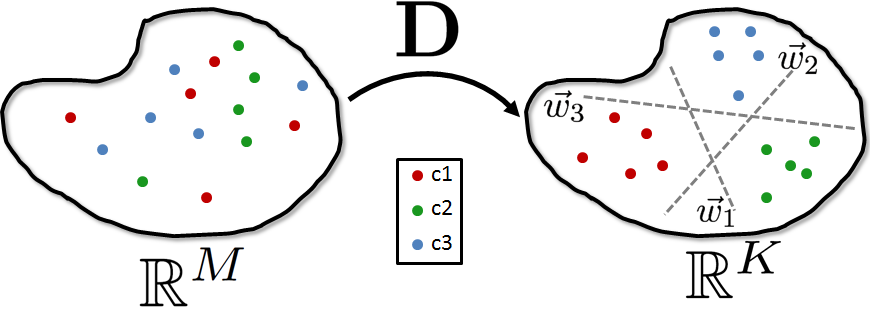}
\caption{An overview of the DDL framework} \label{fig: DDL}
\end{figure}

\else
\iffalse
\begin{wrapfigure}{r}{0.46\textwidth}
\centerline{\includegraphics[width=0.46\textwidth]{pics/rep_class_2.png}}
\caption{An overview of the DDL framework} \label{fig: DDL}
\end{wrapfigure}
\fi
\fi


Although \cite{Mairal2008c,discriminative_reconstructive_hinge} use multiple dictionaries, it is clear that learning a single dictionary allows for sharing of features among labeled classes, less computational cost, and less risk of overfitting. As a result, our proposed method learns a single dictionary $\mathbf{D}$. Here, we note that \cite{LDASparseClassification2006} addresses a similar problem, where $\mathbf{D}$ is predefined and $e_C$ is the Fisher criterion. Despite their merits, SDL methods have the following drawbacks. \textbf{(i)} Most methods use limited forms for $e_C$ (e.g. softmax applied to reconstruction error). Consequently, they cannot generalize to incorporate popular classification costs, such as the exponential loss used in Adaboost or the hinge loss in SVMs. \textbf{(ii)} Previous SDL methods weight the training samples and the classifiers uniformly by setting the fixed mixing coefficient $\lambda$ according to cross-validation. This biases their cost functions to samples that are badly represented or misclassified. As such, they are more sensitive to outlier, noisy, and mislabeled training data. \textbf{(iii)} From an optimization viewpoint, the SDL objective functions are quite involved especially due to the use of the softmax function for multi-class discrimination. 


\iffalse
\textbf{Contributions:} Our proposed DDL framework, illustrated in \figlabel\ref{fig: DDL}, overcomes the previous drawbacks by learning a linear map $\mathbf{D}$ that allows for maximal class discrimination in the labeled data when using linear classification. \textbf{(i)} We prove that this framework is applicable to a general family of classification cost functions, including those used in popular boosting methods. \textbf{(ii)} Since we pose DDL in a probabilistic setting, the representation-classification tradeoff and the weighting of training samples correspond to MAP parameters that are estimated in a data-driven fashion, thus, avoiding parameter tuning. \textbf{(iii)} Since we decouple $e_R$ and $e_C$, the representations $\mathbf{X}$ act as the only liaisons between classification and representation. In fact, this is why well-studied methods in dictionary learning and TSC are only required to solve the DDL problem. This avoids involved optimization techniques. Our framework is efficient, general, and modular, so that any improvement or theoretical guarantee on individual modules (i.e.  TSC or dictionary learning) can be seamlessly incorporated.
\else
\vspace{-2mm}\paragraph{Contributions:} Our proposed DDL framework addresses the previous issues by learning a linear map $\mathbf{D}$ that allows for maximal class discrimination in the labeled data when using linear classification. \textbf{(i)} We show that this framework is applicable to a general family of classification cost functions, including those used in popular boosting methods. \textbf{(ii)} Since we pose DDL in a probabilistic setting, the representation-classification tradeoff and the weighting of training samples correspond to MAP parameters that are estimated in a data-driven fashion that avoids parameter tuning. \textbf{(iii)} Since we decouple $e_R$ and $e_C$, the representations $\mathbf{X}$ act as the only liaisons between classification and representation. In fact, this is why well-studied methods in dictionary learning and TSC can be easily incorporated in solving the DDL problem. This avoids involved optimization techniques. Our framework is efficient, general, and modular, so that any improvement or theoretical guarantee on individual modules (i.e.  TSC or dictionary learning) can be seamlessly incorporated.
\fi

\iftrue
The paper is organized as follows. In Section \ref{sec: overview of models}, we describe the probabilistic representation and classification models in our DDL framework and how they are combined in a MAP setting. Section \ref{sec: learning methodolgy} presents the learning methodology that estimates the MAP parameters and shows how inference is done. In Section \ref{sec: experimental results}, we validate our framework by applying it to digit classification and face recognition and showing that it achieves state-of-the-art performance on benchmark datasets.
\fi

\section{Overview of DDL Framework}\label{sec: overview of models}
In this section, we give a detailed description of the probabilistic models used for representation and classification. Our optimization framework, formulated in a standard MAP setup, seeks to maximize the likelihood of the given labeled data coupled with priors on the model parameters.

\subsection{Representation and Classification Models}
\iftrue
We assume that each $M$-dimensional data sample can be represented as a sparse linear combination of $K$ dictionary atoms with additive Gaussian noise of diagonal covariance: $\vec{\mathbf{y}}=\mathbf{D}\vec{\mathbf{x}}+\vec{\mathbf{n}};~~\vec{\mathbf{n}}\sim\mathcal{N}(\vec{\mathbf{0}},\sigma^{2}\mathbf{I})$. Here, we view the sparse representation $\vec{\mathbf{x}}$ as a latent variable of the representation model. In training, we assume that the training samples are represented by this model. However, test samples can be contaminated by various types of noise that need not be zero-mean Gaussian in nature. In testing, we have: $\vec{\mathbf{y}}=\mathbf{D}\vec{\mathbf{x}}+\vec{\mathbf{e}}+\vec{\mathbf{n}}$, where we constrain any auxiliary noise $\vec{\mathbf{e}}$ (e.g. occlusion) to be sparse in nature without modeling its explicit distribution. This constraint is used in the error correction method for sparse representation in \cite{Wright2009b}. It is clear that the representation in testing is identical to the one in training with the dictionary in the latter being augmented by identity. In both cases, the likelihood of observing a specific $\vec{\mathbf{y}}$ is modeled as a Gaussian: $\left(\vec{\mathbf{y}}|\vec{\mathbf{x}},\mathbf{D}\right)\sim\mathcal{N}\left(\mathbf{D}\vec{\mathbf{x}},\sigma^{2}\mathbf{I}\right)$. Since a single dictionary is used to represent samples belonging to different classes, sharing of features is allowed among classes, which simplifies the learning process.

\fi

To model the classification process, we assume that each data sample corresponds to a label vector $\vec{l}\in\{-1,+1\}^{C}$, which encodes the class membership of this sample, where $C$ is the total number of classes. In our experiments, only one value in $\vec{l}$ is $+1$. We apply a linear classifier (or equivalently a set of additively boosted linear classifiers) to the sparse representations in a one-vs-all classification setup. The probabilistic classification model is shown in Eq. (\ref{eq: classification model}), where $\Omega(.)$ is the classification cost function. Note that appending $1$ to $\vec{\mathbf{x}}$ intrinsically adds a bias term to each classifier $\vec{\mathbf{w}}$. Due to the linearity of the classifier, discrimination of the $j^{\text{th}}$ class is completely determined by the scalar cost function $\Omega\left(\vec{\mathbf{x}}\right)=\Omega\left(z_j\right)$, where $z_j=l_j\vec{\mathbf{w}}_j^T\vec{\mathbf{x}}$. This function quantifies the cost of assigning label $l_j$ to representation $\vec{\mathbf{x}}$ using the $j^{\text{th}}$ classifier $\vec{\mathbf{w}}_j$. For now, we do not specify the functional form of $\Omega(.)$. In Section \ref{sec: learning methodolgy}, we show that most forms of $\Omega(.)$ used in practice are easily incorporated into our DDL framework. Since we seek effective class discrimination, we expect low classification cost for the given representations. Therefore, by arranging all $C$ linear classifiers in matrix $\mathbf{W}$, the event $(\vec{l}|\vec{\mathbf{x}},\mathbf{W})$ can be modeled as a product of $C$ independent exponential distributions parameterized by $\gamma_j$ for $j=1,\ldots,C$. By denoting $\vec{\mathbf{w}}_j$ as the classifier of the $j^{\text{th}}$ class, we have:

\begin{align}
p\left(\vec{l}|\vec{\mathbf{x}},\mathbf{W}\right) \varpropto \frac{1}{\prod_{j=1}^{C}\gamma_j}e^{-\sum_{j=1}^C\frac{1}{\gamma_j}\Omega\left(l_j\vec{\mathbf{w}}_j^T\vec{\mathbf{x}}\right)} \label{eq: classification model}
\end{align}

\subsection{Overall Probabilistic Model}
To formalize notation, we consider a training set of $N$ data samples in $\mathbb{R}^{d}$ that are columns of the data matrix $\mathbf{Y}\in\mathbb{R}^{M\times N}$. The $i^{\text{th}}$ column of the label matrix $\mathbf{L}\in\{+1,-1\}^{C\times N}$ is the label vector $\vec{l}_i$ corresponding to the $i^{\text{th}}$ data sample. Here, we assume that there are $K$ atoms in the dictionary $\mathbf{D}\in \mathbb{R}^{d \times K}$, where $K$ is a fixed integer that is application-dependent. Typically, $K \gg d$. Note that there have been recent attempts to determine an optimal $K$ for a given dataset \cite{Mazhar2008}. For our experiments, $K$ is kept fixed and its optimization is left for future work. The representation matrix $\mathbf{X}\in\mathbb{R}^{K \times N}$ is a sparse matrix, whose columns represent the sparse codes of the data samples $\mathbf{Y}$ using dictionary $\mathbf{D}$. The linear classifiers are columns in matrix $\mathbf{W}\in\mathbb{R}^{K\times C}$. We denote $\Theta_R=\{\sigma_i^2\}_{i=1}^N$ and $\Theta_C=\{\gamma_j\}_{j=1}^C$ as the representation and classification parameters respectively.

In what follows, we combine the representation and classification models from the previous section in a unified framework that will allow for the joint MAP estimation of the unknowns: $\mathbf{D}$, $\mathbf{X}$, $\mathbf{W}$, $\Theta_R$, and $\Theta_C$. By making the standard assumption that the posterior probability consists of a dominant peak, we determine the required MAP estimates by maximizing the product: $p(\mathbf{Y}|\mathbf{D},\mathbf{X},\Theta_R)$$p(\mathbf{L}|\mathbf{X},\mathbf{W},\Theta_C)$$p(\Theta_R)p(\Theta_C)$. Here, we make a simplifying assumption that the prior of the dictionary and representations are uniform. To model the priors of $\Theta_R$ and $\Theta_C$ and to avoid using hyper-parameters, we choose the objective non-parametric Jeffereys prior, which has been shown to perform well for classification and regression tasks \cite{Figueiredo2002}. Therefore, we obtain $p(\Theta_R)\varpropto \prod_{i=1}^N \frac{1}{\sigma_i^2}$ and $p(\Theta_C)\varpropto \prod_{j=1}^C \frac{1}{\gamma_j}$. The motivations behind the selection of these priors are that \textbf{(i)} the representation prior encourages a low variance representation (i.e. the training data should properly fit the proposed representation model) and that \textbf{(ii)} the classification prior encourages a low mean (and variance)\footnote{The mean and variance of an exponential distribution with parameter $\lambda=\frac{1}{\gamma}$ are $\gamma$ and $\gamma^2$ respectively.} classification cost (i.e. the training data should be properly classified using the proposed classification model). By minimizing the sum of the negative log likelihood of the data and labels as well as the log priors, MAP estimation requires solving the optimization problem in Eq. (\ref{eq: total general cost function}), where $\mathbf{L}_{ji}$ represents the label of the $i^{\text{th}}$ training sample with respect to the $j^{\text{th}}$ class.

To encode the sparse representation model, we explicitly enforce sparsity on $\mathbf{X}$ by requiring that each representation $\vec{\mathbf{x}}_i\in\mathcal{S}_T=\{\vec{\mathbf{a}}:\|\vec{\mathbf{a}}\|_0\leq T\}$. An alternative for obtaining sparse representations is to assume that $\vec{\mathbf{x}}_i$ follows a Laplacian prior, which leads to an $\ell_1$ regularizer in the objective. While this sparsifying regularizer alleviates some of the complexity of Eq. (\ref{eq: total general cost function}), it leads to the problem of selecting proper parameters for these Laplacian priors. Note that recent efforts have been made to find optimal estimates of these Laplacian parameters in the context of sparse coding \cite{Giryes2009,Zou2006,Bickel2008}. However, to avoid additional parameters, we choose the form in Eq. (\ref{eq: total general cost function}), where the first two terms of the objective correspond to the representation cost and the last two to the classification cost.

\iffalse
\begin{align}
&\min_{\{\mathbf{D},\mathbf{W},\mathbf{X}\in\mathcal{S}_T^N,\Theta_R,\Theta_C\}} \sum_{i=1}^N\frac{\|\vec{\mathbf{y}}_i-\mathbf{D}\vec{\mathbf{x}}_i\|_2^2}{2\sigma_i^2}+\frac{M+2}{2}\sum_{i=1}^N\ln(\sigma_i^2)\notag\\
&+\sum_{j=1}^C\frac{1}{\gamma_j}\sum_{i=1}^N\Omega\left(\mathbf{L}_{ji}\vec{\mathbf{w}}_j^T\vec{\mathbf{x}}_i\right)+(N+1)\sum_{j=1}^{C}\ln\gamma_j \label{eq: total general cost function}
\end{align}
\else
\begin{align}
&\min_{\{\mathbf{D},\mathbf{W},\mathbf{X},\Theta_R,\Theta_C\}} \sum_{i=1}^N\frac{\|\vec{\mathbf{y}}_i-\mathbf{D}\vec{\mathbf{x}}_i\|_2^2}{2\sigma_i^2}+\sum_{i=1}^N\ln\sigma_i^{M+2}+ \sum_{j=1}^C\sum_{i=1}^N\frac{\Omega\left(\mathbf{L}_{ji}\vec{\mathbf{w}}_j^T\vec{\mathbf{x}}_i\right)}{\gamma_j} +\sum_{j=1}^{C}\ln\gamma_j^{N+1} \label{eq: total general cost function}
\end{align}
\fi

In the following section, we show that Eq. (\ref{eq: total general cost function}) can be solved for a general family of cost functions $\Omega(.)$ using well-known and well-studied techniques in TSC and dictionary learning. In other words, developing specialized optimization methods and performing parameter tuning are \emph{not} required.

\section{Learning Methodology}\label{sec: learning methodolgy}
Since the objective function and sparsity constraints in Eq. (\ref{eq: total general cost function}) are non-convex, we decouple the dependent variables by resorting to a blockwise coordinate descent method (alternating optimization). At each iteration, only a subset of variables is updated at a time. Clearly, learning $\mathbf{D}$ is decoupled from learning $\mathbf{W}$, if $\mathbf{X}$ and $(\Theta_R,\Theta_C)$ are fixed. Next, we identify the four basic update procedures in our DDL framework. In what follows, we  denote the estimate of variable $\mathbf{A}$ at iteration $k$ as $\mathbf{A}^{(k)}$.

\begin{wrapfigure}{r}{0.5\textwidth}
\centering
\subfigure[classification cost]{\includegraphics[width=\ImWidthCostPlots\textwidth]{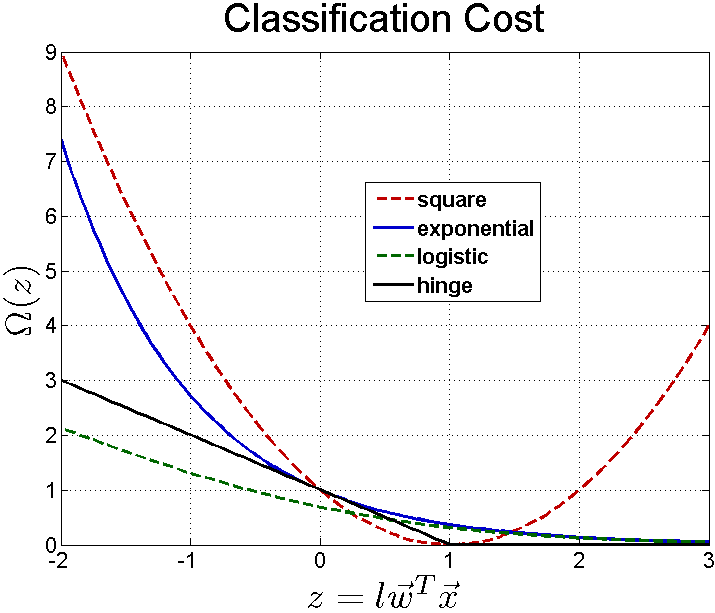}\label{subfig: classification costs}}
\subfigure[classifier weights]{\includegraphics[width=\ImWidthCostPlots\textwidth]{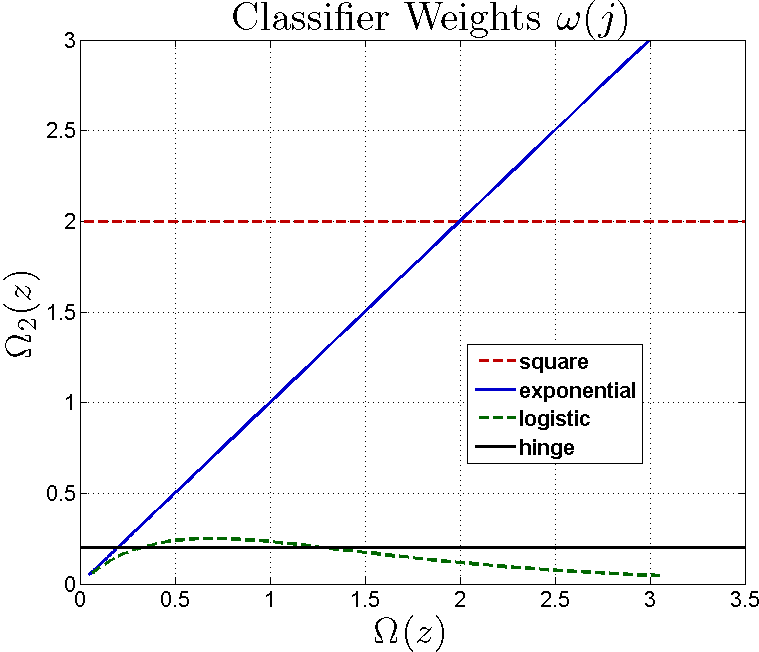}\label{subfig: classifier weights}}
\caption{Four classification cost functions: square, exponential, logistic, and hinge loss in \ref{subfig: classification costs}. \ref{subfig: classifier weights} plots their impacts on classifier weights (second derivatives) in our DDL framework.} \label{fig: DDL classification functions}
\end{wrapfigure}

\subsection{Classifier Update}\label{subsec: classification update}
Since the classification terms in Eq. (\ref{eq: total general cost function}) are decoupled from the representation terms and independent of each other, each classifier can be learned separately. In this paper, we focus on four popular forms of $\Omega(.)$, as shown in \figlabel\ref{subfig: classification costs}: \textbf{(i)} the square loss: $\Omega(z)=(1-z)^2$ optimized by the boosted square leverage method \cite{Duffy2002}, \textbf{(ii)} the exponential loss: $\Omega(z)=e^{-z}$ optimized by the AdaBoost method \cite{Friedman2000a}, \textbf{(iii)} the logistic loss: $\Omega(z)=\ln(1+e^{-z})$ optimized by the LogitBoost method \cite{Friedman2000a}, and \textbf{(iv)} the hinge loss: $\Omega(z)=\max(0,1-z)$ optimized by the SVM method. Since additive boosting of linear classifiers yields a linear classifier, we allow for seamless incorporation of additive boosting, which is a novel contribution.

\subsection{Discriminative Sparse Coding}\label{subsec: discriminative sparse coding}
In this section, we describe how well-known and well-studied TSC algorithms (e.g. Orthogonal Matching Pursuit (OMP)) are used to update $\mathbf{X}^{(k+1)}$ from $\mathbf{X}^{(k)}$. This is done by solving the problem in Eq. (\ref{eq: update representation general}), which we refer to as \emph{discriminative sparse coding} (DSC). DSC requires the sparse code to not only reliably represent the data sample but also to be discriminable by the one-vs-all classifiers. Here, we denote $\vec{l}$ as the label vector of the $i^{\text{th}}$ data element (i.e. the $i^{\text{th}}$ column of $\mathbf{L}$). The $(k)$ superscripts are omitted from variables not being updated to facilitate readability. Here, we note that DSC, as defined here, is a generalization of the functional form used in \cite{LDASparseClassification2006}.

\begin{align}
\vec{\mathbf{x}}_i^{(k+1)}=\arg\min_{\vec{\mathbf{x}}\in\mathcal{S}_T} \left\|\vec{\mathbf{b}}-\mathbf{A}\vec{\mathbf{x}}\right\|_2^2+2\sum_{j=1}^C \frac{\Omega\left(\vec{\mathbf{g}}_j^T\vec{\mathbf{x}}\right)}{\gamma_j}  ~~~~\left[\text{ where } \vec{\mathbf{b}}=\frac{\vec{\mathbf{y}}_i}{\sigma_i}; \mathbf{A}=\frac{\mathbf{D}}{\sigma_i}; \vec{\mathbf{g}}_j= l_j\vec{\mathbf{w}}_j\right]\label{eq: update representation general}
\end{align}

\paragraph{Solving Eq. (\ref{eq: update representation general}):} The complexity of this solution depends on the nature of $\Omega(.)$. However, it is easy to show that, by applying a projected Newton gradient descent method to Eq. (\ref{eq: update representation general}), DSC can be formulated as a sequence of TSC problems, if $\Omega(z)$ is strictly convex. At each Newton iteration, a quadratic local approximation of the cost function is minimized. If we denote $\Omega_1(z)$ and $\Omega_2(z)$ as the first and second derivatives of $\Omega(z)$ respectively and $\Omega_{12}(z)=\frac{\Omega_1(z)}{\Omega_2(z)}$, the quadratic approximation of $\Omega(z)$ around $z_p$ is $\Omega(z)\approx \Omega(z_p)+\Omega_1(z_p)(z-z_p)+\frac{1}{2}\Omega_2(z_p)(z-z_p)^2$. Since $\Omega_2(z)$ is a strictly positive function, we can complete the square to get $\Omega(z)\approx \frac{1}{2}\Omega_2(z_p)[z-(z_p-\Omega_{12}(z_p))]^2+cte$. By replacing this approximation in Eq. (\ref{eq: update representation general}), the objective function at the $(p+1)^{\text{th}}$ Newton iteration is: $\|\vec{\mathbf{b}}-\mathbf{A}\vec{\mathbf{x}}\|_2^2+ \|\mathbf{H}^{(p)}(\vec{\mathbf{\delta}}^{(p)}-\mathbf{G}^T\vec{\mathbf{x}})\|_2^2$. In fact, this objective takes the form of a TSC problem and, thus, can be solved by any TSC algorithm. Here, $\mathbf{G}$ is formed by the columnwise concatenation of $\vec{\mathbf{g}}_j$ and we define $\delta^{(p)}(j)=\vec{\mathbf{g}}_j^T\vec{\mathbf{x}}^{(p)}-\Omega_{12}(\vec{\mathbf{g}}_j^T\vec{\mathbf{x}}^{(p)})$ for $j=1,\ldots,C$. Also, we define the diagonal weight matrix $\mathbf{H}^{(p)}$, where $\mathbf{H}^{(p)}(j,j)=(\frac{\Omega_2(\vec{\mathbf{g}}_j^T\vec{\mathbf{x}}^{(p)})}{2\gamma_j})^{\frac{1}{2}}$ weights the $j^{\text{th}}$ classifier. Based on this derivation, the same TSC algorithm (e.g. OMP) can be used to solve the DSC problem iteratively, as illustrated in Algorithm \ref{algo: DSC}. The convergence of this algorithm is dependent on whether the TSC algorithm is capable of recovering the sparsest solution at each iteration. Although this is not guaranteed in general, the convergence of TSC algorithms to the sparsest solution has been shown to hold, when the solution is sparse enough even if the dictionary atoms are highly correlated \cite{Davenport2010,Wright2009b,Gribonval2004,Donoho2003}. In our experiments, we see that the DSC objective is reduced sequentially and convergence is obtained in almost all cases. Furthermore, we provide a \emph{Stop Criterion} (threshold on the relative change in solution) for the premature termination of Algorithm \ref{algo: DSC} to avoid needless computation.

\begin{algorithm}[htb]
   \caption{Discriminative Sparse Coding (DSC)}\label{algo: DSC}
\begin{algorithmic}
   \STATE {\textbf{INPUT}: $\textbf{A}$, $\vec{\mathbf{b}}$, $\mathbf{G}$, $\vec{\mathbf{\alpha}}$, $\Omega$, $\vec{\mathbf{x}}^{(0)}$, $T$, $p_{\max}$, Stop Criterion}
   \WHILE{~(Stop Criterion) \textbf{AND} $p\leq p_{\max}$}
   \STATE {compute and form: $\vec{\mathbf{\delta}}^{(p)}$ and $\mathbf{H}^{(p)}$;}
   \STATE {$\vec{\mathbf{x}}^{(p+1)}=\text{TSC}\left(\left[\begin{array}{c}\vec{\mathbf{b}}\\ \mathbf{H}^{(p)}\vec{\mathbf{\delta}}^{(p)}\end{array}\right], \left[\begin{array}{c}\mathbf{A}\\ \mathbf{H}^{(p)}\mathbf{G}^T\end{array}\right], T\right);~p=p+1;$}
   \ENDWHILE
   \STATE {\bfseries OUTPUT: $\vec{\mathbf{x}}^{(p)}$}
\end{algorithmic}
\end{algorithm}

\paragraph{Popular Forms of $\Omega(z)$:} Here, we focus on particular forms of $\Omega(z)$, namely the four functions in Section \ref{subsec: classification update}. Before proceeding, we need to replace the traditional hinge cost with a strictly convex approximation. We use the smooth hinge approximation introduced by \cite{Loeff2008b}, which can arbitrarily approximate the traditional hinge. As seen before, $\Omega_{2}(z)$ and $\Omega_{12}(z)$ are the only functions that play a role in the DSC solution. Obviously, only one iteration of Algorithm \ref{algo: DSC} is needed when the square cost is used, since it is already quadratic. For all other $\Omega(z)$, at the $p^{\text{th}}$ iteration of DSC, the impact of the $j^{\text{th}}$ classifier on the overall cost (or equivalently on updating the sparse code) is determined by $\mathbf{H}^{(p)}(j,j)$. This weight is influenced by two terms. \textbf{(i)} It is inversely proportional to $\gamma_j$. So, a classifier with a smaller mean training cost (i.e. higher \emph{training set discriminability}) yields more impact on the solution. \textbf{(ii)} It is proportional to $\Omega_2(l_j\vec{\mathbf{w}}_j^T\vec{\mathbf{x}}^{(p)})$, the second derivative at the previous solution. In this case, the impact of the $j^{\text{th}}$ classifier is determined by the type of classification cost used. In \figlabel\ref{subfig: classifier weights}, we plot the relationship between $\Omega(z)$ and $\Omega_2(z)$ for all four $\Omega(z)$ types. For the square and hinge functions, $\Omega(z)$ and $\Omega_2(z)$ are independent, thus, a classifier yielding high \emph{sample discriminability} (low $\Omega(z)$) is weighted the same as one yielding low discriminability. For the exponential case, the relationship is linear and positively correlated, thus, the lower a classifier's sample discriminability is the higher its weight. This implies that the sparse code will be updated to correct for classifiers that misclassified the training sample in the previous iteration. Clearly, this makes representation sensitive to samples that are ``hard" to classify as well as outliers. This sensitivity is overcome when the logistic cost is used. Here, the relationship is positively correlated for moderate costs but negatively correlated for high costs. This is consistent with the theoretical argument that LogitBoost should outperform AdaBoost when training data is noisy or mislabeled.

\subsection{Unsupervised Dictionary Learning}\label{subsec: unsupervised dictionary learning}
When $\mathbf{X}^{(k)}$, $\Theta_R^{(k)}$, and $\Theta_C^{(k)}$ are fixed, $\mathbf{D}^{(k)}$ can be updated by any unsupervised dictionary learning method. In our experiments, we use the KSVD algorithm, since it avoids expensive matrix inversion operations required by other methods. Also, efficient versions of KSVD have recently been developed \cite{Rubinstein2008}. By alternating between TSC and dictionary updates (SVD operations), KSVD iteratively reduces the overall representation cost and generates a dictionary with normalized atoms and the corresponding sparse representations. In our case, the representations are known apriori, so only a single iteration of the KSVD algorithm is required. For more details, we refer the readers to \cite{KSVDalgo}.

\subsection{Parameter Estimation and Initialization}\label{subsec: parameter update}
The use of the Jeffereys prior for $\Theta_R$ and $\Theta_C$ yields simple update equations:  $\sigma_i^{(k)}=(\frac{1}{M+2}\|\vec{\mathbf{y}}_i-\mathbf{D}^{(k)}\vec{\mathbf{x}}_i^{(k)}\|_2^2)^{\frac{1}{2}}$ and $\gamma_j^{(k)} = \frac{1}{N+1}\sum_{i=1}^N \Omega(\mathbf{L}_{ji}(\vec{\mathbf{w}}_j^{(k)})^T\vec{\mathbf{x}}_i^{(k)})$. These variables estimate the sample representation variance and the mean/variance of the classification cost respectively. Since the overall update scheme is iterative, proper initialization is needed. In our experiments, we initialize $\mathbf{D}^{(0)}$ to a randomly selected subset of training samples (uniformly chosen from the different classes) or to random zero-mean Gaussian vectors, followed by columnwise normalization. Interestingly, both schemes produce similar dictionaries, although the randomized scheme requires more iterations for convergence. The representations $\mathbf{X}^{(0)}$ are computed by TSC using $\mathbf{D}^{(0)}$. Initializing the remaining variables uses the update schemes above. Algorithm \ref{algo: DDL} summarizes the overall DDL framework.

\begin{algorithm}[htb]
   \caption{Discriminative Dictionary Learning (DDL)}\label{algo: DDL}
\begin{algorithmic}
   \STATE {\textbf{INPUT}: $\textbf{Y}$, $\textbf{L}$, $T$, $\Omega$, $q_{\max}$, $p_{\max}$, Stop Criterion}
   \STATE {Initialize $\mathbf{D}^{(0)}$, $\mathbf{X}^{(0)}$, $\Theta_R^{(0)}$, $\Theta_C^{(0)}$, and $q=0$}
   \WHILE{~(Stop Criterion) \textbf{AND} $q\leq q_{\max}$}
   \FOR{$i=1$ \textbf{to} $N$}
   \STATE {$\vec{\mathbf{x}}^{(q+1)}=\text{DSC}($ $\frac{\mathbf{D}^{(q)}}{\sigma_i^{(q)}}$, $\frac{\vec{\mathbf{y}}_i}{\sigma_i^{(q)}}$, $\mathbf{W}^{(q)}\text{diag}(\vec{l}_i)$, $\frac{1}{\Theta_C^{(q)}}$, $\Omega$, $\vec{\mathbf{x}}^{(q)}$, $T$, $p_{\max}$, Stop Criterion$)$;}
   \ENDFOR
   \STATE {Learn classifiers $\mathbf{W}^{(q+1)}$ using $\mathbf{L}$ and $\mathbf{X}^{(q+1)}$;}
   \STATE {$\mathbf{D}^{(q+1)}=\text{KSVD}(\mathbf{D}^{(q)},\mathbf{X}^{(q+1)},T)$;}
   \STATE {Update $\vec{\mathbf{\sigma}}^{(q+1)}$ and $\vec{\mathbf{\gamma}}^{(q+1)}$; $q=q+1$;}
   \ENDWHILE
   \STATE {\bfseries OUTPUT: $\mathbf{D}^{(q)}$, $\mathbf{W}^{(q)}$, $\mathbf{X}^{(q)}$, $\vec{\mathbf{\sigma}}^{(q)}$, and $\vec{\mathbf{\gamma}}^{(q)}$}
\end{algorithmic}
\end{algorithm}

\subsection{Inference}\label{subsec: inference}
After learning $\mathbf{D}$ and $\mathbf{W}$, we describe how the label of a test sample $\vec{\mathbf{y}}_{t}$ is inferred. We seek the class $j_t$ that maximizes $p(\vec{\mathbf{y}}_t|\vec{l}_t(j))$, where $\vec{l}_t(j)$ is the label vector of $\vec{\mathbf{y}}_t$ assuming it belongs to class $j$. By marginalizing with respect to $\vec{\mathbf{x}}$ and assuming a single dominant representation $\vec{\mathbf{x}}_t$ exists, $j_t$ is the class that maximizes $p(\vec{\mathbf{y}}_t|\vec{\mathbf{x}}_t,\mathbf{D})p(\vec{\mathbf{x}}_t|\vec{l}_t(j),\mathbf{W})$, as in Eq. (\ref{eq: inference of class label}). The inner maximization problem is exactly a DSC problem where $\vec{l}_t(j)$ is the hypothesized label vector. Here, we use the testing representation model to account for dense errors (e.g. occlusion), thus, augmenting $\mathbf{D}$ by identity. Computing $j_t$ involves $C$ independent DSC problems. To reduce computational cost, we solve a single TSC problem instead: $\vec{\mathbf{x}}_t=\argmax_{\vec{\mathbf{x}}\in S_T} p(\vec{\mathbf{y}}_t|\vec{\mathbf{x}},\mathbf{D})$. In this case, $j_t=\argmax_{j\in{1,\ldots,C}} p(\vec{l}_t(j)|\vec{\mathbf{x}}_t,\mathbf{W})$.

\begin{align}
j_t=\argmax_{j\in{1,\ldots,C}} \left(\max_{\vec{\mathbf{x}}\in S_T}p(\vec{\mathbf{y}}_t|\vec{\mathbf{x}},\mathbf{D})p(\vec{l}_t(j)|\vec{\mathbf{x}},\mathbf{W})\right) \label{eq: inference of class label}
\end{align}

\paragraph{Implementation Details:}
There are several ways to speedup computation and allow for quicker convergence. \textbf{(i)} The DSC update step is the most computationally expensive operation in Algorithm \ref{algo: DDL}. This is mitigated by using a greedy TSC method (Batch-OMP instead of $\ell_1$ minimization methods) and exploiting the inherent parallelism of DDL (e.g. doing DSC updates in parallel). \textbf{(ii)} Selecting suitable initializations for $\mathbf{D}$ and the DSC solutions can dramatically speedup convergence. For example, choosing $\mathbf{D}^{(0)}$ from the training set leads to a smaller number of DDL iterations than randomly choosing $\mathbf{D}^{(0)}$. Also, we initialize DSC solutions at a given DDL iteration with those from the previous iteration. Moreover, the DDL framework is easily extended to the semi-supervised case, where only a subset of training samples are labeled. The only modification to be made here is to use TSC (instead of DSC) to update the representations of unlabeled samples.

\section{Experimental Results}\label{sec: experimental results}
In this section, we provide empirical analysis of our DDL framework when applied to handwritten digit classification ($C=10$) and face recognition ($C=38$). Digit classification is a standard machine learning task with two popular benchmarks, the USPS and MNIST datasets. The digit samples in these two datasets have been acquired under different conditions or written using significantly different handwriting styles. To alleviate this problem, we use the alignment and error correction technique for TSC that was introduced in \cite{RobustFaceRecognition}. This corrects for gross errors that might occur (e.g. due to thickening of handwritten strokes or reasonable rotation/translation). Consequently, we do not need to augment the training set with shifted versions of the training images, as done in \cite{TaskDrivenDL2010}. Furthermore, we apply DDL to face recognition, which is a machine vision problem where sparse representation has made a big impact. We use the Extended Yale B (E-YALE-B) benchmark for evaluation. To show that learning $\mathbf{D}$ in a discriminative fashion improves upon traditional dictionary learning, we compare our method against a baseline that treats representation and classification independently. In the baseline, $\mathbf{X}$ and $\mathbf{D}$ are estimated using KSVD, $\mathbf{W}$ is learned using $\mathbf{X}$ and $\mathbf{L}$ directly, and a a winner-take-all classification strategy is used. Clearly, our framework is general, so we do not expect to outperform methods that use domain-specific features and machinery. However, we \emph{do} achieve results comparable to state-of-the-art. Also, we show that our DDL framework significantly outperforms the baseline. In all our experiments, we set $q_{\text{max}}=20$ and $p_{\text{max}}=100$ and initialize $\mathbf{D}$ to elements in the training set.

\begin{figure}
\begin{narrow}{0mm}{0mm}
\centering
$\begin{array}{cc}
\includegraphics[width=\ImWidthErrorPlots\columnwidth]{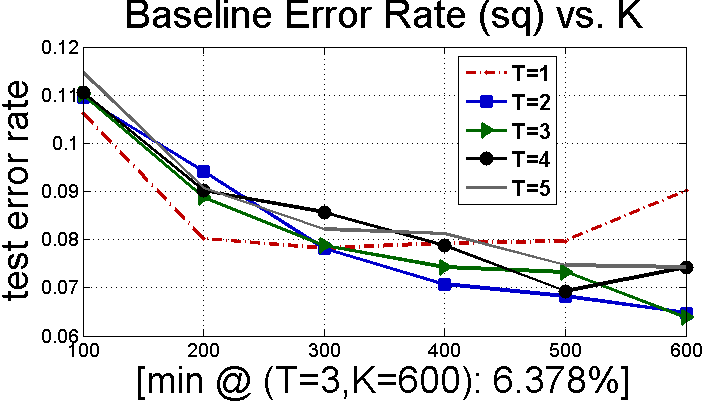} & \includegraphics[width=\ImWidthErrorPlots\columnwidth]{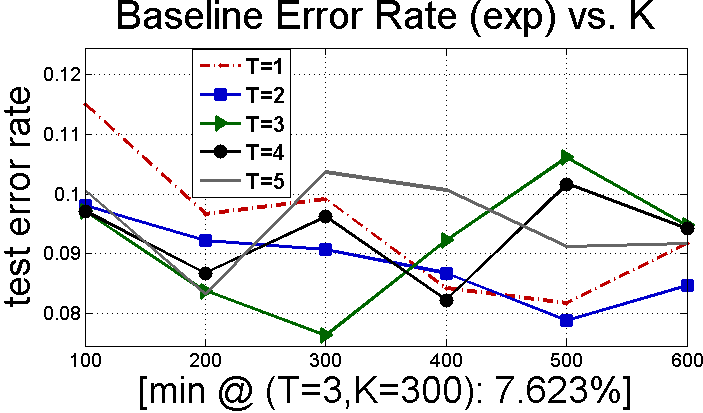}\\
\includegraphics[width=\ImWidthErrorPlots\columnwidth]{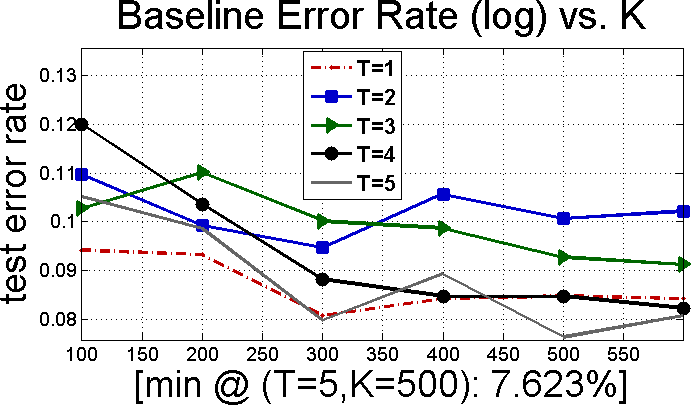} & \includegraphics[width=\ImWidthErrorPlots\columnwidth]{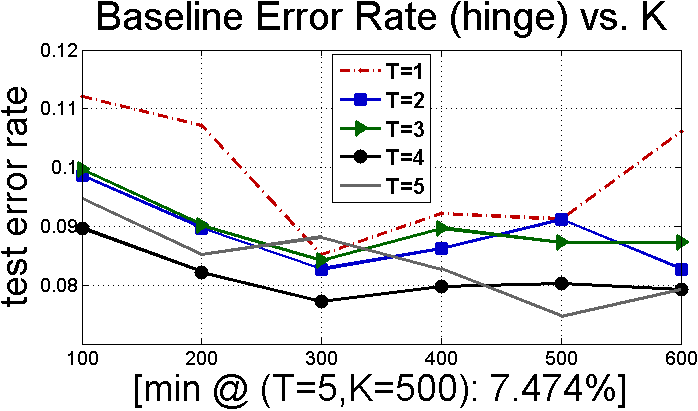}
\end{array}$
\end{narrow}
\caption{Baseline classification performance on the USPS dataset} \label{fig: baseline performance USPS}
\end{figure}

\begin{figure}
\begin{narrow}{0mm}{0mm}
\centering
$\begin{array}{cc}
\includegraphics[width=\ImWidthErrorPlotss\columnwidth]{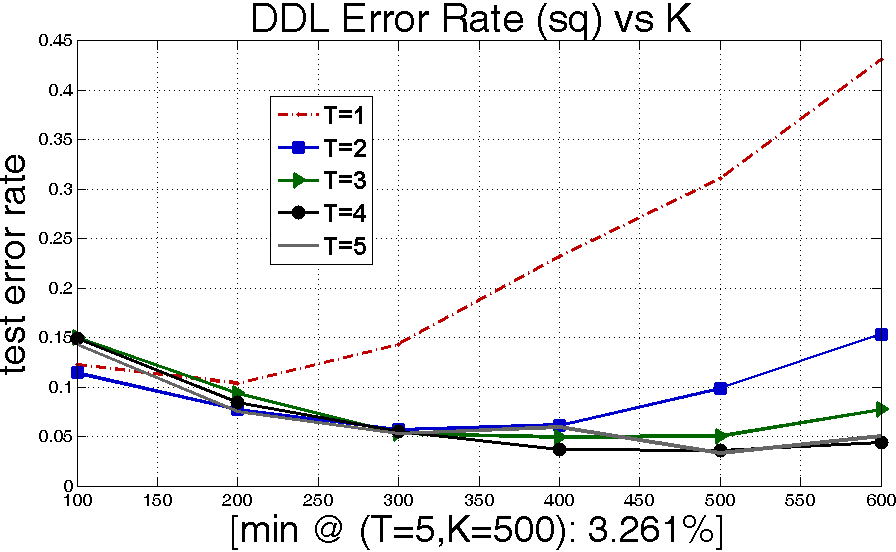} & \includegraphics[width=\ImWidthErrorPlotss\columnwidth]{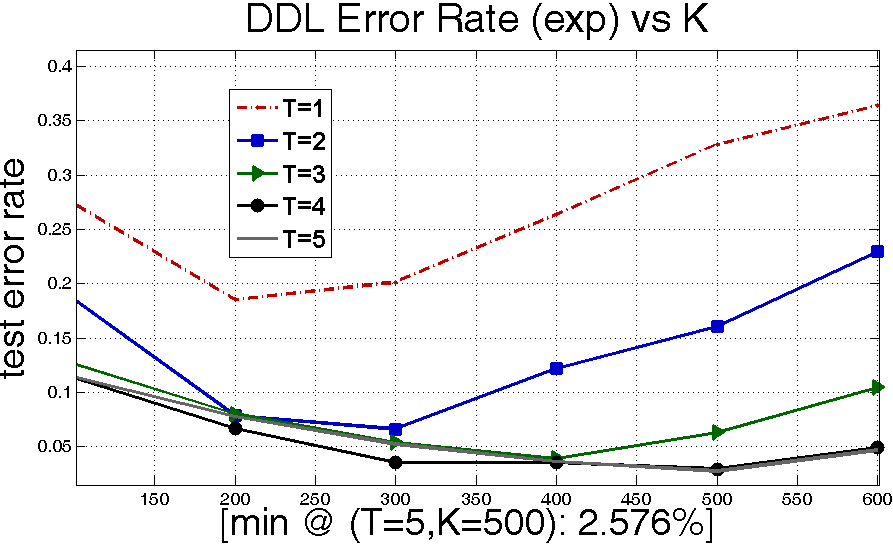}\\
\includegraphics[width=\ImWidthErrorPlotss\columnwidth]{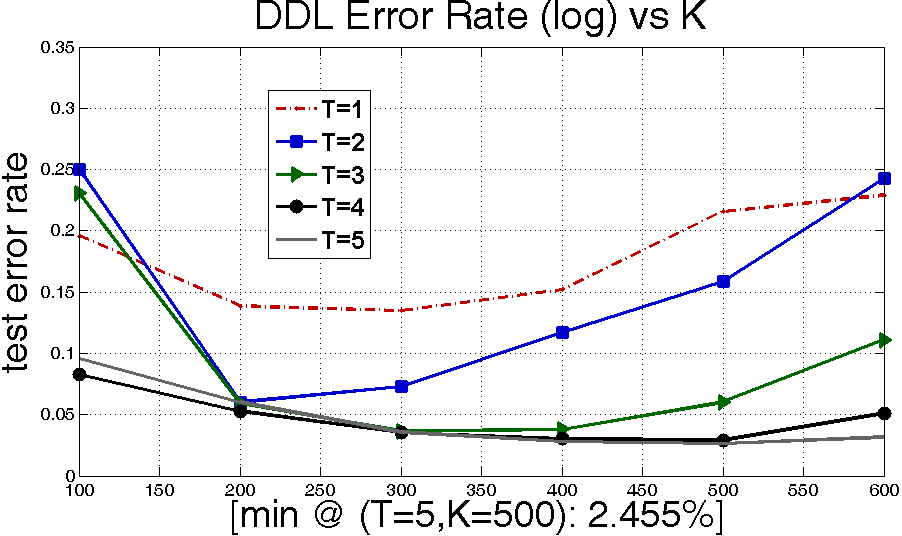} & \includegraphics[width=\ImWidthErrorPlotss\columnwidth]{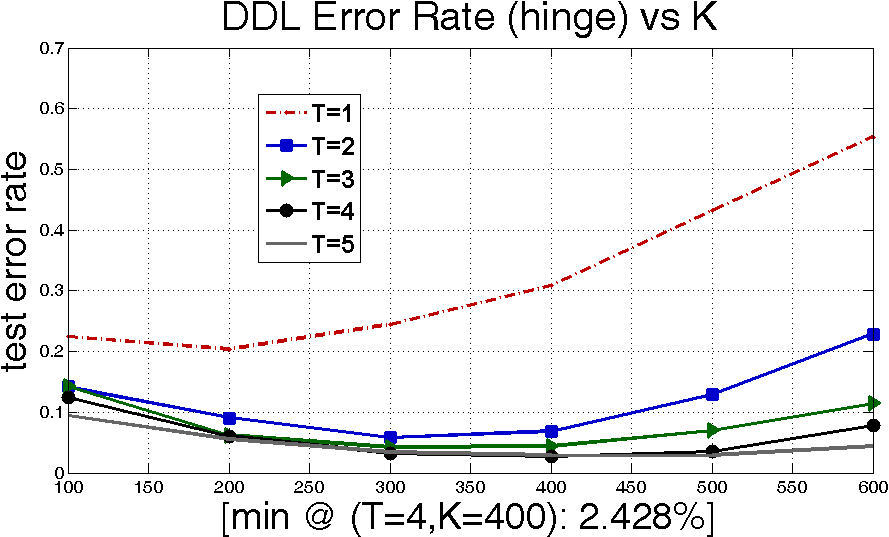}
\end{array}$
\end{narrow}
\caption{DDL classification performance on the USPS dataset} \label{fig: our performance USPS}
\end{figure}

\paragraph{Digit Classification:} The USPS dataset comprises $N=7291$ training and $2007$ test images, each of $16\times 16$ pixels ($M=256$). We plot the test error rates of the baseline for the four classifier types and for a range of $T$ and $K$ values in \figlabel\ref{fig: baseline performance USPS}. Beneath each plot, we indicate the values of $K$ and $T$ that yield minimum error. This is a common way of reporting SDL results \cite{TaskDrivenDL2010,SDLNIPS2008,discriminative_reconstructive_hinge,Mairal2008c}. Interestingly, the square loss classifier leads to the lowest error and the best generalization. For comparison, we plot the results of our DDL method in \figlabel\ref{fig: our performance USPS}. Clearly, our method achieves a significant improvement of $4.5\%$ over the baseline, and $1\%$ and $0.5\%$ over the SDL methods in \cite{SDLNIPS2008} and \cite{TaskDrivenDL2010} respectively. Our results are comparable to the state-of-the-art performance ($2.2\%$) \cite{Keysers2000}). This result shows that adapting $\mathbf{D}$ to the underlying data \emph{and} class labels yields a dictionary that is better suited for classification. Increasing $T$ leads to an overall improvement of performance because representation becomes more reliable. However, we observe that beyond $T=3$, this improvement is insignificant. The square loss classifier achieves the lowest performance and the logistic classifier achieves the highest. The variations of error with $K$ are similar for all the classifiers. Error steadily decreases till an ``optimal" $K$ value is reached. Beyond this $K$ value, performance deteriorates due to overfitting. Future work will study how to automatically predict this optimal value from training data, without resorting to cross-validation.

\begin{figure}
\begin{narrow}{0mm}{0mm}
\centering
$\begin{array}{cc}
\includegraphics[width=\ImWidthParamPlots\columnwidth]{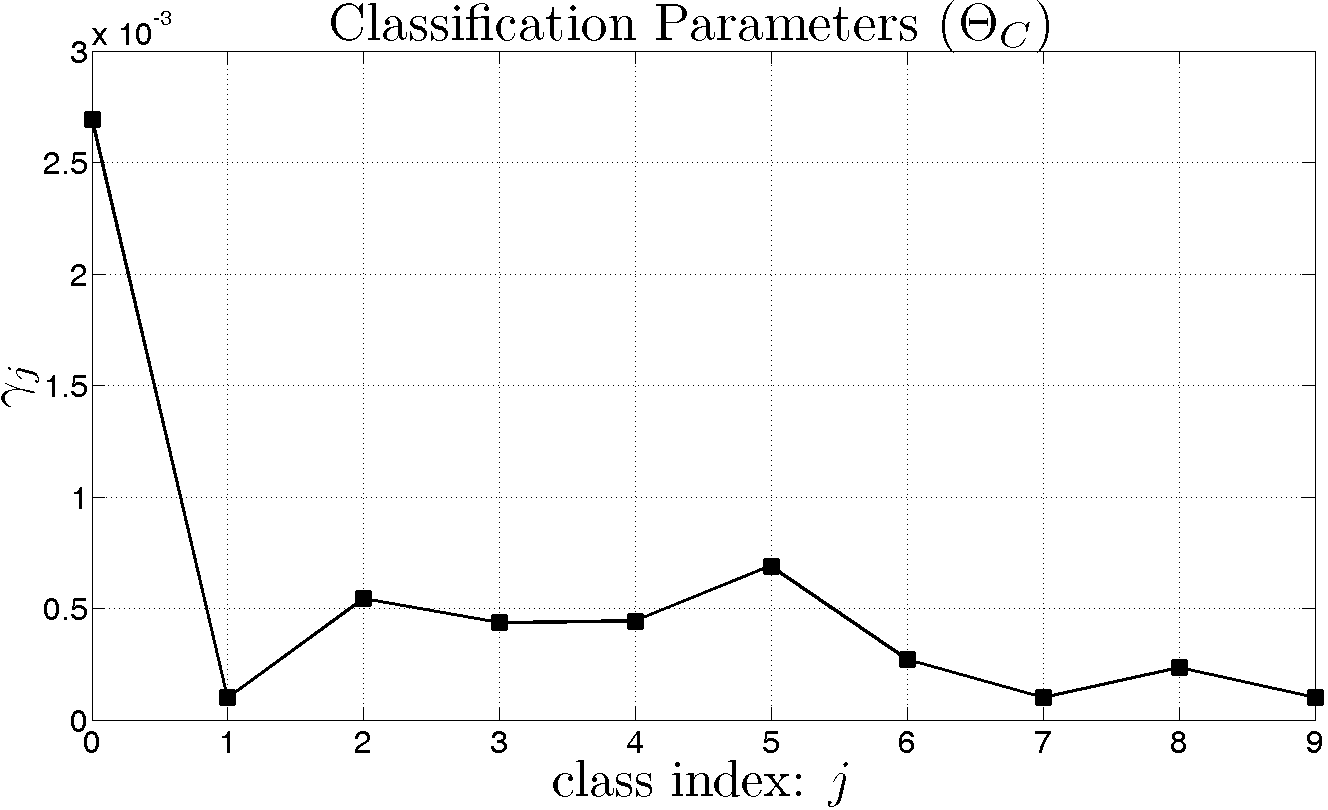} & \includegraphics[width=\ImWidthParamPlotss\columnwidth]{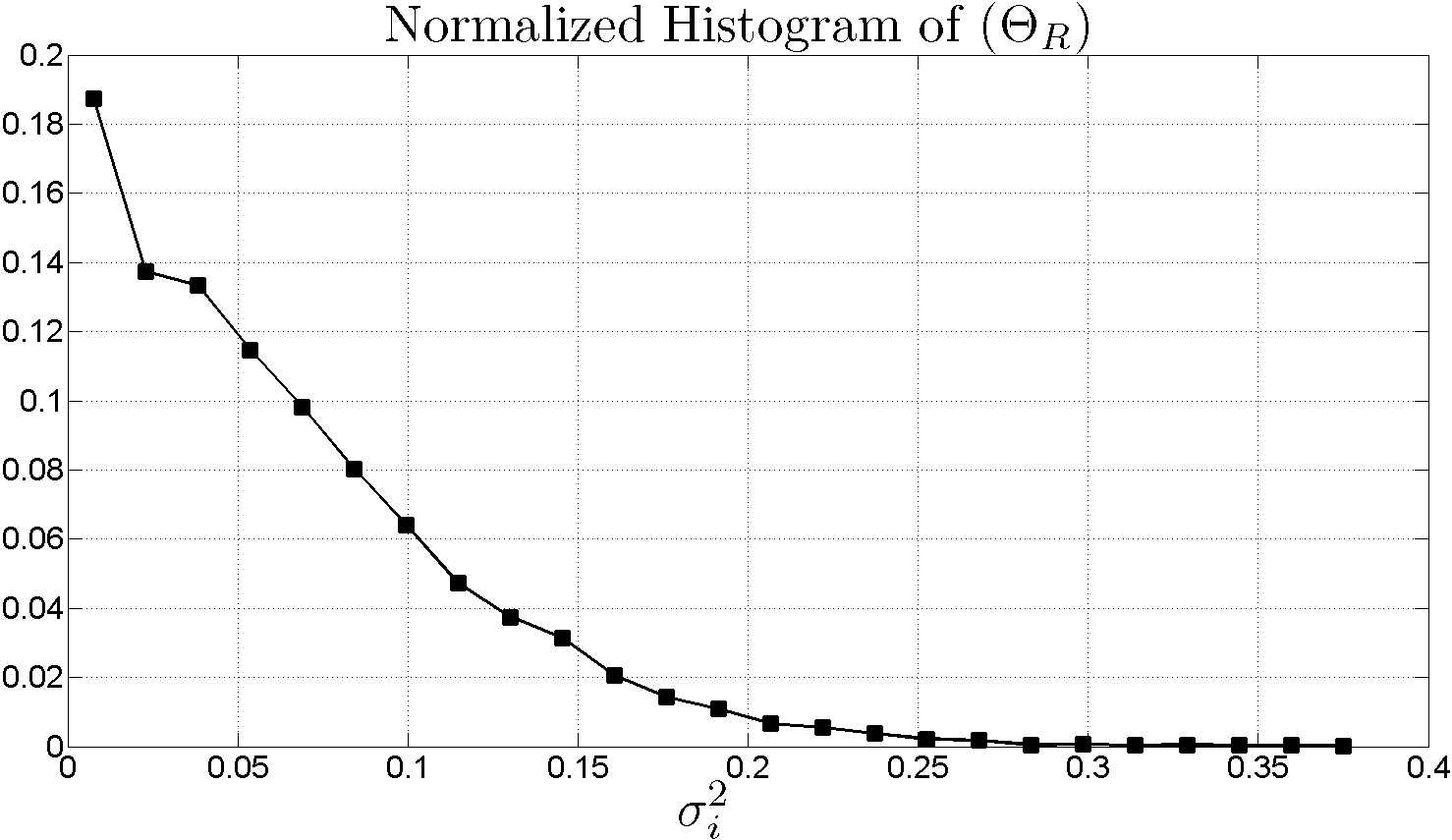}
\end{array}$
\end{narrow}
\caption{Parameters $\Theta_R$ and $\Theta_C$ learned from the USPS dataset}\label{fig: DDL parameters}
\end{figure}

In \figlabel \ref{fig: DDL parameters}, we plot the learned parameters $\Theta_R$ (in histogram form) and $\Theta_C$ for a typical DDL setup. We observe that the form of these plots does not significantly change when the training setting is changed. We notice that the $\Theta_R$ histogram fits the form of the Jeffereys prior, $p(x)\varpropto \frac{1}{x}$. Most of the $\sigma$ values are close to zero, which indicates reliable reconstruction of the data. On the other hand, $\Theta_C$ take on similar values for most classes, except the ``0" digit class that contains a significant amount of variation and thus the highest classification cost. Note that these values tend to be inversely proportional to the classification performance of their corresponding linear classifiers. We provide a visualization of the learned $\mathbf{D}$ in the \textbf{supplementary material}. Interestingly, we observe that the dictionary atoms resemble digits in the training set and that the number of atoms that resemble a particular class is inversely proportional to the accuracy of that class's binary classifier. This occurs because a ``hard" class contains more intra-class variations requiring more atoms for representation.

The MNIST dataset comprises $N=60000$ training and $10000$ test images, each of $28\times28$ pixels ($M=784$). We show the baseline and DDL test error rates in Table \ref{table: MNIST YALE performance}. We train each classifier type using the $K$ and $T$ values that achieved minimum error for that classifier on the USPS dataset. Compared to the baseline, we observe a similar improvement in performance as in the USPS case. Also, our results are comparable to state-of-the-art performance ($0.53\%$) for this dataset \cite{Jarrett2009}.

\paragraph{Face Recognition:} The E-YALE-B dataset comprises $2,414$ images of $C=38$ individuals, each of $192\times 168$ pixels, which we downsample by an order of $8$ ($M=504$). Using a classification setup similar to \cite{Wright2009} with $K=600$ and $T=5$, we record the classification results in Table \ref{table: MNIST YALE performance}, which lead to implications similar to those in our previous experiments. Interestingly, DDL achieves similar results to the robust sparse representation method of \cite{Wright2009}, which uses all training samples ($K\approx1200$) as atoms in $\mathbf{D}$. This shows that learning a discriminative $\mathbf{D}$ can reduce the dictionary size by as much as $50\%$, without significant loss in performance.

\begin{table}[htb]
\caption{Baseline and DDL test error on MNIST and E-YALE-B datasets}\label{table: MNIST YALE performance}
\begin{center}
\begin{tabular}{l|cccc||cccc|}
\cline{2-9}
& \multicolumn{4}{|c||}{MNIST (digit classification)} & \multicolumn{4}{c|}{E-YALE-B (face recognition)} \\ \cline{2-9}
& SQ    & EXP   & LOG     & HINGE & SQ    & EXP   & LOG    & HINGE \\ \cline{1-9}
\multicolumn{1}{|c|}{BASELINE} & $8.35\%$ & $6.91\%$ & $5.77\%$ & $4.92\%$
& $10.23\%$ & $9.65\%$ & $9.23\%$ & $9.17\%$\\

\multicolumn{1}{|c|}{DDL} & \textbf{1.41}$\%$ & \textbf{1.28}$\%$ & \textbf{1.01}$\%$ & \textbf{0.72}$\%$
& \textbf{8.89}$\%$ & \textbf{7.82}$\%$ & \textbf{7.57}$\%$ & \textbf{7.30}$\%$\\
\hline
\end{tabular}
\end{center}
\end{table}

\section{Conclusions}\label{sec: conclusions}
This paper addresses the problem of discriminative dictionary learning by jointly learning a sparse linear representation model and a linear classification model in a MAP setting. We develop an optimization framework that is capable of incorporating a diverse family of popular classification cost functions and solvable by a sequence of update operations that build on well-known and well-studied methods in sparse representation and dictionary learning. Experiments on standard datasets show that this framework outperforms the baseline and achieves state-of-the-art performance.

\normalsize{\bibliography{supervised_dict_learning_nips_refs}
\bibliographystyle{ieee}}

\end{document}